\documentclass{article}
\usepackage{spconf,amsmath,amssymb,graphicx,booktabs,array,placeins,stfloats,flushend,hyperref}
\hypersetup{hidelinks}
\newcommand{\Real}{\textup{REAL SOFT}}

\newcommand{\DeepTailCE}{\textup{DeepTailCE}}

\title{Cluster Assignments in Soft Targets Shape Speech Representations: Evidence from S-JEPA}
\name{Wenxuan He$^{1}$ \quad Yunpeng Li$^{1}$ \quad Zewei Li$^{1}$ \quad Yongke Yang$^{1}$ \quad Yuze Li$^{1}$ \quad Yin Cao$^{2}$ \quad Shan Liang$^{1,*}$}
\address{{\normalsize $^{1}$ Academy of Artificial Intelligence and Advanced Technology, Xi'an Jiaotong-Liverpool University, Suzhou, China}\\[-2pt]
{\normalsize $^{2}$ Institute of Acoustics, Chinese Academy of Sciences, Beijing, China; $^{*}$ Corresponding author}}
\begin{document}
\ninept
\maketitle

\begin{abstract}
Cluster-based prediction is widely used in self-supervised speech learning. A soft target preserves a distribution over clusters rather than a single label. This distribution specifies both the probability values and which clusters receive them. Comparisons between soft targets and hard labels do not separate the contributions of these two aspects to the learned representation. We study this in S-JEPA, a recent high-performing self-supervised speech model trained with soft Gaussian mixture model (GMM) targets. We compare its original targets with counterfactual targets that preserve the most likely cluster and all probability values but change which remaining clusters receive the other probabilities. Across three training seeds, the original soft distribution is recovered more accurately from Encoders trained with the original than counterfactual targets. Because this could reflect target matching alone, we also test low-level acoustic and phonetic information. Both are more accessible from Encoders trained with the original targets. This suggests that cluster assignments affect acoustic and phonetic properties of the learned representation, not just recovery of the training target.
\end{abstract}

\section{Introduction}
\label{sec:intro}

Cluster-based prediction is an important approach to self-supervised speech learning~\cite{hsu2021hubert,chen2022wavlm,chiu2022bestrq,liu2023dinosr,ma2023units}. Models such as HuBERT and WavLM use hard cluster labels as training targets, assigning each speech frame to a single cluster~\cite{hsu2021hubert,chen2022wavlm}. Soft targets instead retain a probability distribution over clusters~\cite{ioannides2026gmmjepa}. S-JEPA, a recent high-performing model trained with soft Gaussian mixture model (GMM) targets, demonstrates that this approach can support strong speech representations~\cite{ioannides2026sjepa}. Prior benchmarks have evaluated speech representations across downstream tasks~\cite{yang2021superb,yang2024large}. Representation analyses have examined the acoustic, phonetic, and linguistic information accessible across models and layers~\cite{chung2021similarity,pasad2021layerwise,ashihara2024what,liu2024mutual,pasad2024words,choi2024phonetic,english2022phonetic}. However, how the additional structure retained by soft targets is reflected in the learned representation remains unclear.

\begin{figure*}[t]
  \centering
  \includegraphics[width=.910\textwidth]{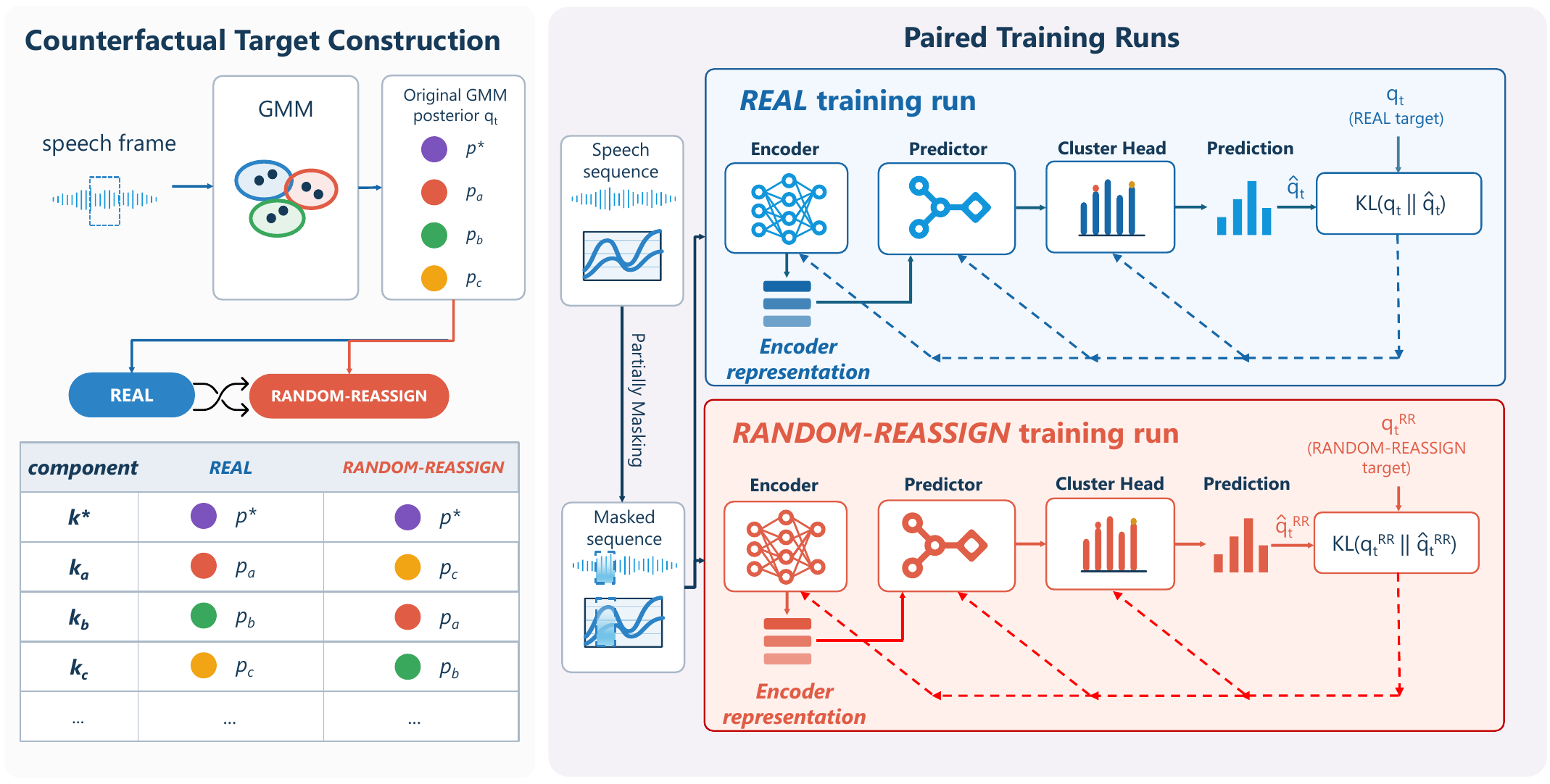}
  \caption{Counterfactual target construction and matched training runs.
  REAL uses the original fixed-GMM posterior $q_t$, whereas RANDOM-REASSIGN
  preserves the top-1 component and all posterior values while reassigning
  only the non-maximal values to form $q_t^{\mathrm{RR}}$.
  The two runs share the masked input, architecture, and training setup,
  and minimize KL divergence to their corresponding targets.
  Encoder outputs are used for subsequent representation analysis;
  solid and dashed arrows denote forward and gradient flow, respectively.}
  \label{fig:framework}
\end{figure*}

A soft target retains both the probability values and their assignments to clusters. Hardening removes the non-maximal values and the information about which clusters received them, so a comparison with hard labels cannot isolate the separate effects of probability values and cluster assignments. Changing these assignments necessarily changes the training target. What remains unclear is whether their effect is limited to target recovery or extends to other properties of the learned representation. S-JEPA, a recent representative model using soft targets for self-supervised speech learning, allows the effect of cluster assignments to be isolated from that of probability values because its initial training stage uses a fixed GMM, so each position in the target distribution refers to the same cluster throughout training~\cite{ioannides2026sjepa}. We compare the original targets with counterfactual targets that preserve the most likely cluster and every probability value but randomly reassign the remaining values among the other clusters, keeping each frame's reassignment fixed throughout training (Fig.~\ref{fig:framework}). The target construction differs only in these assignments, and the training conditions are otherwise matched.

Across three independent training seeds, the original non-maximal distribution is more accurately recovered by linear readouts from frozen Encoders trained with the original targets than from those trained with the counterfactual targets. However, this advantage could simply reflect closer agreement with the target used to train each Encoder. To determine whether the difference extends beyond target recovery, we first test low-level acoustic dynamics defined from changes in the speech spectrum, then phone categories defined from transcripts. The effect therefore extends beyond direct target recovery to acoustic and phonetic information accessible from the learned representation. The same pattern also appears across intermediate levels of retained assignment, rather than only at the two endpoints.

Our study introduces a counterfactual design that isolates cluster assignment from probability values in soft cluster targets. The results show that the effect of these assignments can extend beyond direct target recovery to acoustic and phonetic information in the learned representation.

\section{COUNTERFACTUAL DESIGN AND REPRESENTATION ANALYSIS}
\label{sec:target}

A controlled counterfactual separates cluster assignments from probability values. We then test where any corresponding representation difference is detectable, from direct target recovery to properties defined outside the GMM posterior. Graded exposure examines how the same readouts vary with the fraction of training frames retaining their original assignments.

\subsection{Changing assignments while preserving values}

RANDOM-REASSIGN is a controlled counterfactual that isolates cluster assignments from probability values. It preserves every value in the GMM posterior and reassigns only the non-maximal values. Because the Phase-1 GMM is fixed before training, each posterior position refers to the same component throughout Phase 1~\cite{ioannides2026sjepa}.

Let $q_t\in\Delta^{K-1}$ denote the fixed Phase-1 GMM posterior for frame $t$, where $K=100$. The component with the largest posterior value and that value are
\begin{equation*}
k_t^*=\arg\max_{k\in\{1,\ldots,K\}}q_{t,k},
\qquad
p_t^*=q_{t,k_t^*}.
\end{equation*}
The remaining component indices form
\begin{equation*}
\mathcal{T}_t=\{1,\ldots,K\}\setminus\{k_t^*\}.
\end{equation*}

REAL SOFT uses $q_t$ unchanged. For each frame, RANDOM-REASSIGN samples a permutation $\pi_t$ of $\mathcal{T}_t$ and keeps it fixed throughout Phase 1:
\begin{equation}
q^{\mathrm{RR}}_{t,k}=
\begin{cases}
p_t^*, & k=k_t^*,\\
q_{t,\pi_t(k)}, & k\in\mathcal{T}_t.
\end{cases}
\label{eq:rp}
\end{equation}

Equation~\eqref{eq:rp} preserves the largest posterior value, its corresponding GMM component, and the complete multiset of non-maximal values. Tail mass, entropy, and vector norm are identical. The only change is which remaining GMM component receives each non-maximal value.

\subsection{Representation readouts}
\label{sec:readouts}

The readouts test whether any representation difference associated with cluster assignments is confined to direct target recovery or appears in acoustic and phonetic readouts defined outside the GMM posterior. GMM-tail recovery measures access to the original non-maximal distribution. Controlled spectral dynamics assesses low-level acoustic information from the speech spectrum, and phone classification assesses transcript-defined phonetic information.

\noindent\textbf{GMM-tail recovery.}
For frames with non-maximal mass
\begin{equation*}
m_t=1-p_t^*\geq 0.1,
\end{equation*}
we remove the original maximal entry and normalize the remaining probabilities:
\begin{equation*}
\tilde q_{t,k}=
\frac{q_{t,k}}{1-p_t^*},
\qquad k\in\mathcal T_t.
\end{equation*}
A linear readout predicts $\tilde q_t$ from the frozen Encoder representation. Recovery is measured by cross-entropy, with lower values indicating better recovery.

\noindent\textbf{Controlled spectral dynamics.}
Let $x_t\in\mathbb R^{80}$ denote the normalized log-Mel spectrum, and define
\begin{equation*}
\Delta^2x_t=
\frac{x_{t+2}-2x_t+x_{t-2}}{4}.
\end{equation*}
We remove the part linearly predictable from the complete current-frame spectrum, phone identity, GMM top-1 component, posterior entropy, top-1 probability, and position relative to the nearest phone boundary. A linear readout predicts the resulting residual from the frozen Encoder representation. We report $R^2$, with higher values indicating better prediction.

\noindent\textbf{Phone identity.} Forced alignments provide 39 non-silence ARPAbet classes. We exclude SIL, SP, SPN, and empty labels, while retaining valid non-silence frames at phone boundaries and transitions. A linear classifier predicts phone identity from the frozen Encoder representation, and accuracy measures phonetic accessibility~\cite{martin2023aspiration,mohamed2024orthogonality}.

\subsection{Graded exposure}

Graded exposure examines how each readout varies with the fraction of training frames retaining their original cluster assignments. REAL SOFT and RANDOM-REASSIGN provide only the two endpoints, so we evaluate the same readouts at intermediate levels.

Each frame receives a fixed routing score
\begin{equation*}
u_t\sim\operatorname{Uniform}[0,1),
\end{equation*}
sampled independently of its reassignment. At
\begin{equation*}
\lambda\in\{0,0.25,0.5,0.75,1\},
\end{equation*}
frames with $u_t<\lambda$ use REAL SOFT, and the remaining frames use RANDOM-REASSIGN. The same routing scores are used across all levels, giving nested REAL SOFT subsets and a fixed target for every frame within each training run.

\section{Experimental Protocol}
\label{sec:protocol}

\subsection{Model, data, and matched Phase-1 training}

We use the S-JEPA configuration in~\cite{ioannides2026sjepa}, including a six-layer 768-dimensional Transformer Encoder and a 100-way Phase-1 head trained against GMM posteriors by KL divergence. Phase 1 runs for 10k updates with AdamW~\cite{loshchilov2019adamw}, a learning rate of $10^{-4}$, 500 warmup updates, weight decay of $10^{-3}$, and a global batch size of 64.

Training uses a fixed 20 h LibriSpeech subset with 5,723 utterances from 51 speakers~\cite{panayotov2015librispeech}. Probe settings are selected on dev-clean. Final evaluation uses 2,620 test-clean utterances from 40 disjoint speakers. The fixed 100-component diagonal GMM is fitted only on the Phase-1 training subset from static, delta, and delta-delta MFCCs~\cite{davis1980mfcc,furui1986dynamic}. We initialize it with MiniBatch $k$-means~\cite{sculley2010minibatch} and fit 20 expectation-maximization iterations~\cite{dempster1977em}.

Within each seed, all runs share initialization, data order, crops, masks, optimizer, and training schedule. Only the training targets differ, as defined in Section~\ref{sec:target}. REAL SOFT and RANDOM-REASSIGN are trained under three independent seeds A--C. For graded exposure, the three intermediate levels $\lambda\in\{0.25,0.5,0.75\}$ are trained under the same seeds. At $\lambda=0$ and $\lambda=1$, the targets reduce to RANDOM-REASSIGN and REAL SOFT, respectively. All five exposure levels are evaluated with the same three readouts.

\subsection{Readout implementation}

At each checkpoint, the Encoder is frozen and separate probes are fitted on dev-clean representations from L4, L5, and L6. They are evaluated on test-clean, and the reported score is the mean over the three layers. The Predictor, cluster head, and GMM are not used as prediction modules.

For GMM-tail recovery, a multi-output Ridge model with $\alpha=1$ predicts the target defined in Section~\ref{sec:readouts}. Its outputs are clipped to nonnegative values, zeroed at the original top-1 entry, and renormalized before cross-entropy is computed. A floor of $10^{-8}$ is used for normalization and the logarithm. The resulting cross-entropy is denoted DeepTailCE.

For controlled spectral dynamics, the log-Mel sequence uses utterance-level mean and variance normalization. Reflection is applied at utterance boundaries. Phone labels and boundaries come from forced alignment of the official LibriSpeech transcripts. The covariate Ridge model is fitted on dev-clean. A second Ridge model predicts the residual target from each Encoder layer. We report pooled test $R^2$, averaged over L4--L6.

Phone identity is evaluated with a linear multinomial softmax classifier applied to one frame at a time~\cite{niekerk2021speaker,liu2023orthogonal}. A separate classifier is fitted at each of L4, L5, and L6 using $C=0.01$ for 100 epochs. DeepPhoneAcc is the mean test accuracy over L4--L6.

\subsection{Phase-2 continuation and statistical analysis}

For the primary seed (Seed A), we continue the final REAL SOFT and RANDOM-REASSIGN checkpoints separately with S-JEPA's original Phase-2 procedure~\cite{ioannides2026sjepa} after reassignment is removed. Each branch maintains its own EMA Encoder and a separate 500-component online GMM built from its evolving L5 representation. Subsequent soft targets are therefore generated from each branch's evolving representation rather than altered directly by the Phase-1 intervention. Evaluations are performed at the Phase-1 endpoint and after 10k, 50k, and 100k Phase-2 updates using the same three readouts.

Within each Phase-1 seed, REAL SOFT is compared with RANDOM-REASSIGN. For each seed, we fit an ordinary least squares line with an intercept to the five readout values across the exposure levels. We report the fitted slope and the endpoint contrast between $\lambda=1$ and $\lambda=0$.

For REAL SOFT and RANDOM-REASSIGN comparisons, 95\% confidence intervals use 10,000 paired speaker-cluster bootstrap draws over the 40 test speakers~\cite{davison1997bootstrap}. Each draw samples speakers with replacement and retains all utterances from the selected speakers. The same draw is applied to REAL SOFT and RANDOM-REASSIGN and across the Phase-2 checkpoints. For controlled dynamic $R^2$, the selected speakers' sufficient statistics are pooled before $R^2$ is recomputed. These intervals quantify evaluation-speaker uncertainty with the trained models fixed. Phase-1 seeds A--C are independent training replications. Phase 2 continues Seed A only.

\section{RESULTS}
\label{sec:results}

\begin{figure*}[t]
  \centering
  \includegraphics[width=.96\textwidth]{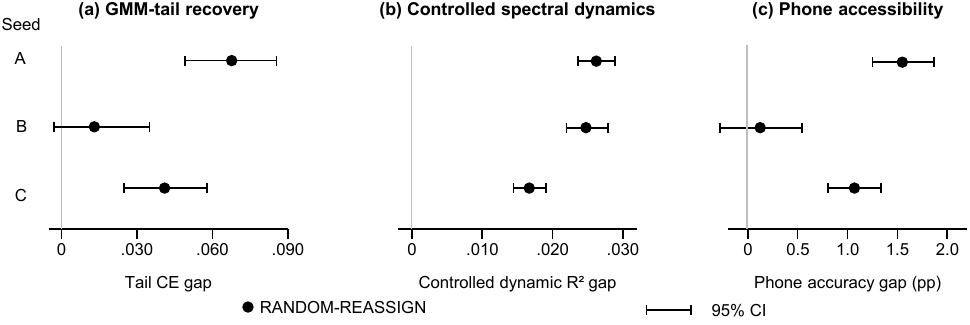}
  \caption{Phase-1 differences between \Real{} and
  \textup{RANDOM-REASSIGN} across three independent seeds. Points show the
  test-clean gap for each seed; error bars are paired 95\% speaker-cluster
  bootstrap confidence intervals over the 40 test speakers (10,000 draws).
  Panel (a) reports RANDOM-REASSIGN minus REAL SOFT in DeepTailCE. Panels (b)
  and (c) report REAL SOFT minus RANDOM-REASSIGN in controlled dynamic $R^2$
  and phone accuracy. Positive values therefore favor REAL SOFT in every
  panel.}
  \label{fig:replication}
\end{figure*}

\subsection{Phase-1 endpoint comparison}

At the two Phase-1 endpoints, we first test how well the original
non-maximal distribution can be recovered from the frozen Encoder.
Figure~\ref{fig:replication}a shows that REAL SOFT lowers DeepTailCE by
0.0679, 0.0135, and 0.0402 in Seeds A--C. Because the two training
targets contain the same posterior values, this difference cannot be
attributed to the values themselves.

However, this readout is itself defined from the original GMM posterior,
so the result does not show whether the difference extends beyond target
recovery. We therefore turn to controlled spectral dynamics and phone
identity (Fig.~\ref{fig:replication}b,c). Controlled dynamic $R^2$ increases
by 0.0262, 0.0247, and
0.0167 across Seeds A--C, while phone accuracy increases by 1.55, 0.12,
and 1.07 percentage points. The assignment effect therefore extends
beyond direct recovery of the training posterior to low-level acoustic
dynamics and phonetic information. Table~\ref{tab:core_effects} summarizes
the Phase-1 endpoint gains across seeds.

\begin{table}[t]
  \centering
  \caption{\Real{} gains over \textup{RANDOM-REASSIGN} across three Phase 1 seeds and after
  100k Phase 2 updates for the primary seed.}
  \label{tab:core_effects}
  \vspace{3pt}
  \small
  \setlength{\tabcolsep}{3.2pt}
  \begin{tabular}{@{}lcccc@{}}
    \toprule
    & \multicolumn{3}{c}{Phase 1} & Phase 2 \\
    \cmidrule(lr){2-4}\cmidrule(l){5-5}
    Readout & A & B & C & \multicolumn{1}{c}{A (100k)} \\
    \midrule
    \DeepTailCE{} reduction & 0.0679 & 0.0135 & 0.0402 & 0.0370 \\
    Controlled dynamic $R^2$ gain & 0.0262 & 0.0247 & 0.0167 & 0.0052 \\
    Phone accuracy gain (pp) & 1.55 & 0.12 & 1.07 & 1.77 \\
    \bottomrule
  \end{tabular}
\end{table}

\subsection{Graded exposure}

\begin{figure*}[t]
  \centering
  \includegraphics[width=.98\textwidth]{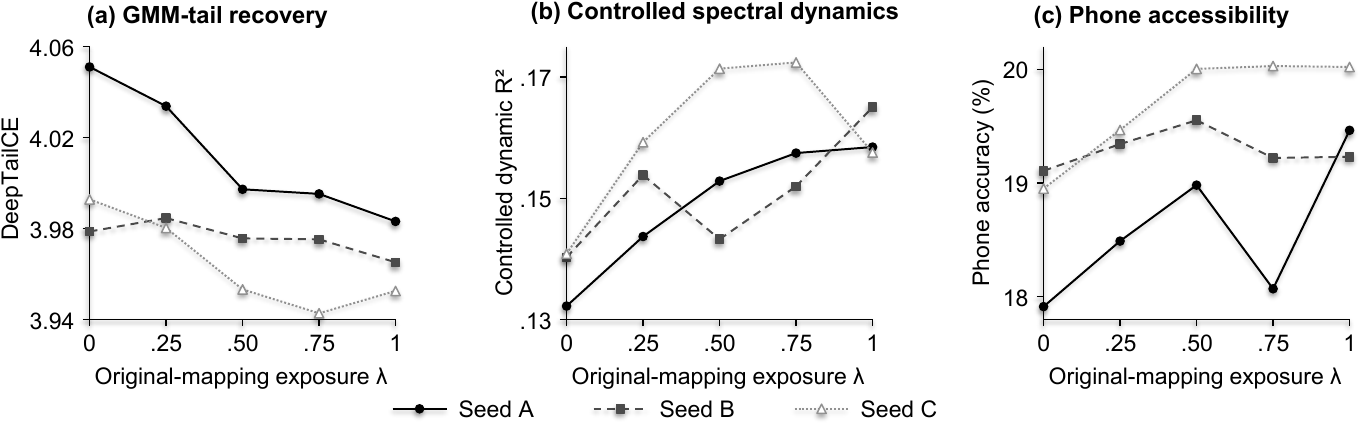}
  \caption{Graded exposure to the original component correspondence. Each line
  shows one independent Phase-1 seed across
  $\lambda\in\{0,0.25,0.5,0.75,1\}$. Panels report (a) DeepTailCE, where lower
  is better; (b) controlled dynamic $R^2$; and (c) phone accuracy, where higher
  is better. At $\lambda=0$, all frames use RANDOM-REASSIGN; at $\lambda=1$,
  all frames use REAL SOFT. Intermediate levels retain the original
  correspondence for nested subsets of frames.}
  \label{fig:graded_exposure}
\end{figure*}

The endpoint comparison uses only fully reassigned and fully original targets.
Graded exposure examines how the three readouts change as more training frames
retain their original assignments. Figure~\ref{fig:graded_exposure}a shows
that DeepTailCE decreases overall as $\lambda$ increases in all three seeds. The
fitted slopes are $-0.0698$, $-0.0146$, and $-0.0472$ for Seeds A--C.
Controlled dynamic $R^2$ increases with slopes of $+0.0265$, $+0.0190$,
and $+0.0186$ (Fig.~\ref{fig:graded_exposure}b). Phone accuracy has slopes
of $+1.07$, $+0.05$, and $+1.08$ percentage points
(Fig.~\ref{fig:graded_exposure}c). The slope directions are consistent across all three
seeds and readouts, giving an overall trend toward the original-target endpoint
rather than only a difference between the two endpoints.

\subsection{Phase-2 continuation}

The Phase-1 experiments establish representation differences under a fixed
target generator. Phase 2 tracks how these differences evolve when S-JEPA
begins generating soft targets from the evolving representations. For the
primary seed, both branches resume S-JEPA's original Phase-2 procedure after
reassignment is removed.

The gaps do not evolve uniformly. The DeepTailCE difference remains positive
at all evaluated Phase-2 checkpoints and is 0.0370 after 100k updates. The controlled-dynamics gap
decreases to 0.0052 over the same period. The phone-accuracy gap remains close
to its Phase-1 magnitude and reaches 1.77 percentage points after 100k updates.
These Phase-2 endpoint values are reported in Table~\ref{tab:core_effects}. In this
primary-seed continuation, the controlled-dynamics difference decreases
substantially under representation-dependent target generation, but the
tail-recovery and phone-accessibility differences remain detectable.

\section{Conclusion}
\label{sec:conclusion}

Our results show that the role of soft cluster targets in self-supervised speech learning cannot be understood from their probability values alone. In S-JEPA, preserving all posterior values while changing the assignments of the non-maximal probabilities affects not only recovery of the original target, but also low-level acoustic dynamics and phone identity accessible from the learned representation. These differences show consistent overall trends as more training frames retain their original assignments, rather than arising only between the two endpoint targets. These results show that, for soft cluster targets, the effect of cluster assignments can extend beyond direct target recovery to acoustic and phonetic information in the learned representation.

\medskip
\noindent\textit{Funding:} This work was supported by the Internal Research Project of Xi'an Jiaotong-Liverpool University (RDF-24-01-016).
\par\noindent\textit{Compliance with Ethical Standards:} This study uses publicly available LibriSpeech data for secondary analysis and involves no new participant recruitment or data collection. No additional ethical approval was required for this secondary analysis. The authors declare no relevant financial or non-financial conflicts of interest.

\bibliographystyle{IEEEbib}
{\fontsize{9}{9.2}\selectfont
\bibliography{refs}
}
\end{document}